\documentclass[]{IEEEtran}
\IEEEoverridecommandlockouts
\usepackage{cite}
\usepackage{amsmath,amssymb,amsfonts}
\usepackage{algorithmic}
\usepackage{graphicx}
\usepackage{textcomp}
\usepackage{xcolor}
\usepackage{times}
\usepackage{comment}
\usepackage{epsfig}
\usepackage{graphicx}
\usepackage{amsmath}
\usepackage{amssymb}
\usepackage{graphicx}
\usepackage{url}
\usepackage{array}
\usepackage{multirow}
\usepackage{subcaption}
\usepackage{color}
\usepackage{resizegather}
\usepackage{makecell}
\usepackage{textcomp}
\usepackage{resizegather}
\usepackage{cite}
\usepackage{footnote}
\def\BibTeX{{\rm B\kern-.05em{\sc i\kern-.025em b}\kern-.08em
    T\kern-.1667em\lower.7ex\hbox{E}\kern-.125emX}}

\usepackage{fancyhdr}
\fancypagestyle{firstpage}{%
  \lfoot{\small{This paper is published by Journal of Visual Communication and Image Representation, Elsevier. The final paper is available at: https://www.sciencedirect.com/science/article/pii/S1047320321002522.}}
}

\begin{document}

\title{CDGAN: Cyclic Discriminative Generative Adversarial Networks for Image-to-Image Transformation}
\author{Kancharagunta Kishan Babu and Shiv Ram Dubey
\thanks{K.K. Babu and S.R. Dubey were with Indian Institute of Information Technology, Sri City, Andhra Pradesh - 517646, India.}
\thanks{S.R. Dubey is now associated with Indian Institute of Information Technology, Allahabad, Uttar Pradesh - 211015, India.
{\tt\small email: srdubey@iiita.ac.in}}
}

\maketitle
\thispagestyle{firstpage}

\begin{abstract}
Generative Adversarial Networks (GANs) have facilitated a new direction to tackle the image-to-image transformation problem. Different GANs use generator and discriminator networks with different losses in the objective function. Still there is a gap to fill in terms of both the quality of the generated images and close to the ground truth images. In this work, we introduce a new Image-to-Image Transformation network named Cyclic Discriminative Generative Adversarial Networks (CDGAN) that fills the above mentioned gaps. The proposed CDGAN generates high quality and more realistic images by incorporating the additional discriminator networks for cycled images in addition to the original architecture of the CycleGAN. The proposed CDGAN is tested over three image-to-image transformation datasets. The quantitative and qualitative results are analyzed and compared with the state-of-the-art methods. The proposed CDGAN method outperforms the state-of-the-art methods when compared over the three baseline Image-to-Image transformation datasets. The code is available at \url{https://github.com/KishanKancharagunta/CDGAN}.
\end{abstract}


\section{Introduction}
There are many real world applications where images of one particular domain need to be translated into different target domain. For example, sketch-photo synthesis \cite{FSSMAL}, \cite{DCFFPSS}, \cite{MRBFSPS} is required in order to generate the photo images from the sketch images that helps to solve many real time law and enforcement cases, where it is difficult to match sketch images with the gallery photo images due to domain disparities. Similar to sketch-photo synthesis, many image processing and computer vision problems need to perform the image-to-image transformation task, such as Image Colorization, where gray-level image is translated into the colored image \cite{DC}, \cite{CIC}, Image in-painting, where lost or deteriorated parts of the image are reconstructed \cite{HRIPMSNPS}, \cite{FLIP}, Image, video and depth map super-resolution, where resolution of the images is enhanced \cite{PLRTSTSR}, \cite{VSR}, Artistic style transfer, where the semantic content of the source image is preserved while the style of the target image is transferred to the source image \cite{DLISR}, \cite{SISRDL}, and Image denoising, where the original image is reconstructed from the noisy measurement \cite{ANLAID}. Some other applications like rain or haze removal from the images \cite{RRNRain}, \cite{idrcgan}, \cite{vhrgan}, deblurring \cite{zhao2020gradient}, Radial Distortion Rectification \cite{dr-gan}, visualization \cite{li2020improved}, \cite{fu2020conditional}, cross-modal representation \cite{zhang2020semi}, generating realistic videos \cite{grvkgan}, predicting skeletal activity for early activity recognition \cite{cui2020ap} are also needed to perform image-to-image transformation. 
However, traditionally the image-to-image transformation methods are proposed for a particular specified task with the specialized method, which is suited for that task only. 

\begin{figure}[t]
\centering
\includegraphics[width=\linewidth , height=4.5 cm]{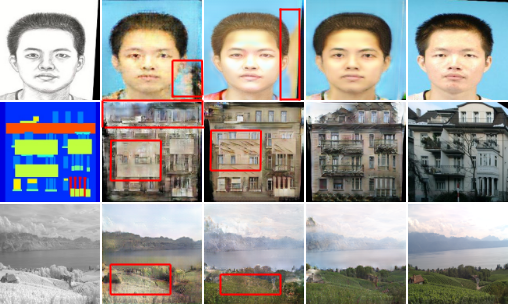}
\caption{A few generated samples obtained from our experiments. The $1^{st}$, $2^{nd}$, and $3^{rd}$ rows represent the Sketch-Photos from CUHK Face Sketch dataset \cite{chuk}, Labels-Buildings from FACADES dataset \cite{facades} and RGB-NIR scenes from RGB-NIR Scene dataset \cite{rgb_nir}, respectively. The $1^{st}$ column represents the input images. The $2^{nd}$, $3^{rd}$ and $4^{th}$ columns show the generated samples using DualGAN \cite{dualGAN}, CycleGAN \cite{CyclicGAN}, and introduced CDGAN methods, respectively. The last column shows the ground truth images. The artifacts generated are highlighted with red color rectangles in $2^{nd}$ and $3^{rd}$ columns corresponding to DualGAN and CycleGAN, respectively.}
\label{fig:qualititative}
\end{figure}

\subsection{CNN Based Image-to-Image Transformation Methods}
Deep Convolutional Neural Networks (CNN) are used as an end-to-end  frameworks for image-to-image transformation problems. The CNN consists of a series of convolutional and deconvolutional layers. It tries to minimize the single objective (loss) function in the training phase while learns the network weights that guide the image-to-image transformation. In the testing phase, the given input image of one visual representation is transformed into another visual representation with the learned weights. For the first time, Long et al. \cite{FCNSS} have shown that the convolutional networks can be trained in end-to-end fashion for pixelwise prediction in semantic segmentation problem. Larsson et al. \cite{AC} have developed a fully automatic colorization method for translating the grayscale images to the color images using a deep convolutional architecture. Zhang et al. \cite{fccn} have proposed a novel method for end-to-end photo-to-sketch synthesis using the fully convolutional network. Gatys et al. \cite{ISTCNN} have introduced a neural algorithm for image style transfer, that constrains the texture synthesis with learned feature representations from the state-of-the art CNNs.
Feng et al. \cite{feng2018dual} have investigated a Dual Swap Disentangling (DSD) weakly semi-supervised method for learning interpretable disentangled representations with the help of dual autoencoder structure.
These methods treat image transformation problem separately and designed CNNs suited for that particular problem only. It opens an active research scope to develop a common framework that can work for the different image-to-image transformation problems.

\begin{figure*}[t]
\begin{center}
\includegraphics[width=\linewidth, height=6cm]{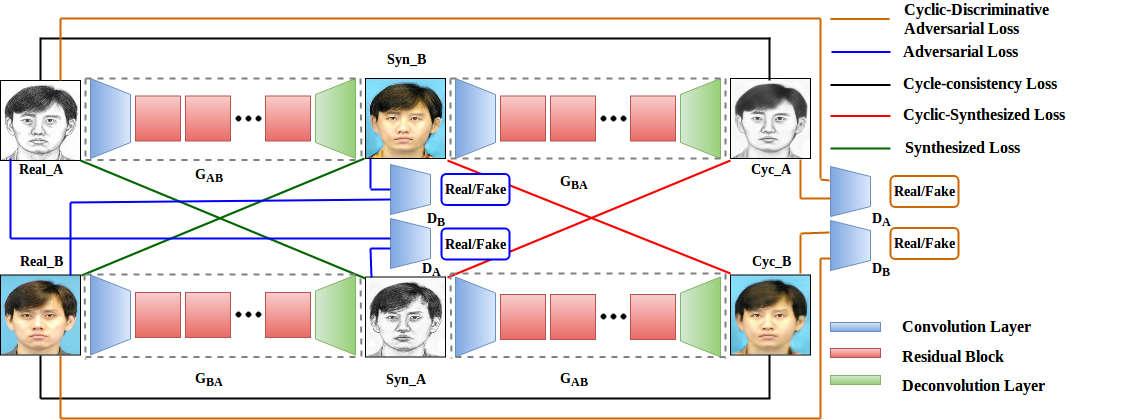}
\caption{Image-to-image transformation framework based on proposed Cyclic Discriminative Generative Adversarial Networks (CDGAN) method. $G_{AB}$ and $G_{BA}$ are the generators, $D_A$ and $D_B$ are the discriminators, $Real_A$ and $Real_B$ are the real images, $Syn_A$ and $Syn_B$ are Synthesized images, $Cyc_A$ and $Cyc_B$ are the Cycled images in domain $A$ and domain $B$, respectively.}
\label{fig_CDGAN_architecture}
\end{center}
\end{figure*}

\subsection{GAN Based Image-to-Image Transformation Methods}
In 2014, Ian Goodfellow et al. \cite{gan} have introduced Generative Adversarial Networks (GAN) as a general purpose solution to the image generation problem. Rather than generating images of the given random noise distribution as mentioned in \cite{gan}, GANs are also used for different computer vision applications like image super resolution \cite{PRSISRGAN}, real-time style transfers \cite{PLRTSTSR}, sketch-to-photo synthesis \cite{FAGSPS}, text-to-image synthesis \cite{bridge-gan}, driver drowsiness detection or recognition \cite{ddr3dcgan} and domain adaptation \cite{RS-DAN}. Mizra et al. \cite{conditionalGAN} have introduced conditional GANs (cGAN) by placing a condition on both of the generator and the discriminator networks of the basic GAN \cite{gan} using class labels as an extra information. Conditional GANs have boosted up the image generation problems. Since then, an active research is being conducted to develop the new GAN models that can work for different image generation problems. Still, there is a need to develop the common methods that can work for different image generation problems, such as sketch-to-face, NIR-to-RGB, etc.

Isola et al. have introduced Pix2Pix \cite{cGAN}, as a general purpose method consisting a common framework for the image-to-image transformation problems  using Conditional GANs (cGANs). Pix2Pix works only for the paired image datasets. It consists the generative adversarial loss and $L_2$ loss as an objective function. Wang et al. have proposed PAN \cite{PAN}, a general framework for image-to-image transformation tasks by introducing the perceptual adversarial loss and combined it with the generative adversarial loss. 
Zhu et al. have investigated CycleGAN \cite{CyclicGAN}, an image-to-image transformation framework that works over unpaired datasets also. The unpaired datasets make it difficult to train the generator network due to the discrepancies in the two domains. This leads to a mode collapse problem where the majority of the generating images share the common properties results in similar images as outputs for different input images. To overcome this problem Cycle-consistency loss is introduced along with the adversarial loss. Yi et al. have developed DualGAN \cite{dualGAN}, a dual learning framework for image-to-image transformation of the unsupervised data. DualGAN consists of the reconstruction loss (similar to the Cycle-consistency loss in \cite{CyclicGAN}) and the adversarial loss as an objective function.
Wang et al. have introduced PS2MAN \cite{ps2-man}, a high quality photo-to-sketch synthesis framework consisting of multiple adversarial networks at different resolutions of the image. PS2MAN consists Synthesized loss in addition to the Cycle-consistency loss and adversarial loss as an objective function. Recently, Kancharagunta et al. have proposed CSGAN \cite{csgan}, an image-to-image transformation method using GANs. CSGAN considers the Cyclic-Synthesized loss along with the other losses mentioned in \cite{CyclicGAN}. 

 Most of the above mentioned methods consist of two generator networks $G_{AB}$ and $G_{BA}$ which read the input data from the Real\_Images ($R_A$ and $R_B$) from the domains $A$ and $B$, respectively. These generators $G_{AB}$ and $G_{BA}$ generate the Synthesized\_Images ($Syn_B$ and $Syn_A$) in domain $B$ and $A$, respectively. The same generator networks $G_{AB}$ and $G_{BA}$ are also used to generate the Cycled\_Images ($Cyc_B$ and $Cyc_A$) in the domains $B$ and $A$ from the Synthesized\_Images $Syn_A$ and $Syn_B$, respectively. In addition to the generator networks $G_{AB}$ and $G_{BA}$, the above mentioned methods also consist of the two discriminator networks $D_A$ and $D_B$. These discriminator networks $D_A$ and $D_B$ are used to distinguish between the Real\_Images ($R_A$ and $R_B$) and the Synthesized\_Images ($Syn_A$ and $Syn_B$) in the its respective domains. The losses, namely Adversarial loss, Cyclic-Consistency loss, Synthesized loss and Cyclic-Synthesized loss are used in the objective function.
 
In this paper, we introduce a new architecture called Cyclic-Discriminative Generative Adversarial Network (CDGAN) for the image-to-image transformation problem. It improves the architectural design by introducing a Cyclic-Discriminative adversarial loss computed among the Real\_Images and the Cycled\_Images. The CDGAN method also consists of two generators $G_{AB}$ and $G_{BA}$ and two discriminators $D_A$ and $D_B$ similar to other methods. The generator networks $G_{AB}$ and $G_{BA}$ are used to generate the Synthesized\_Images ($Syn_B$ and $Syn_A$) from the Real\_Images ($R_A$ and $R_B$) and the Cycled\_Images ($Cyc_B$ and $Cyc_A$) from the Synthesized\_Images ($Syn_A$ and $Syn_B$) in two different domains $A$ and $B$, respectively. The two discriminator networks $D_A$ and $D_B$ are used to distinguish between the Real Images ($R_A$ and $R_B$) and the Synthesized Images ($Syn_A$ and $Syn_B$) and also between the Real Images ($R_A$ and $R_B$) and the Cycled Images ($Cyc_A$ and $Cyc_B$).
 
Following is the main contributions of this work:
\begin{itemize}
  \item We propose a new method called Cyclic Discriminative Generative Adversarial Network (CDGAN), that uses Cyclic-Discriminative (CD) adversarial loss computed over the Real\_Images and the Cycled\_Images. This loss helps to increase the quality of the generated images and also reduces the artifacts in the generated images.
  \item We evaluate the proposed CDGAN method over three benchmark image-to-image transformation datasets with four different benchmark image quality assessment measures.
  \item We conduct the ablation study by extending the concept of proposed Cyclic-Discriminative adversarial loss between the Real\_Images and the Cycled\_Images with the state-of-the art methods like CycleGAN \cite{CyclicGAN}, DualGAN \cite{dualGAN}, PS2GAN \cite{ps2-man} and CSGAN.
\end{itemize}

The remaining paper is arranged as follows; Section \ref{Proposed Method} presents the proposed method and the losses used in the objective function; Experimental setup with the datasets and evaluation metrics used in the experiment are shown in section \ref{Exp Setup}. Result analysis and ablation study are conducted in section \ref{Result Analysis} followed by the Conclusions in section \ref{Conclusion}. 


\section{Proposed CDGAN Method} \label{Proposed Method}
Consider a paired image dataset $X$ between two different domains $A$ and $B$ represented as $X \in \{(A_i),(B_i)\}_{i=1}^{n}$ where $n$ is the number of pairs. The goal of the proposed CDGAN method is to train two generator networks $G_{AB}: A \rightarrow B$ and $G_{BA}: B\rightarrow A $ and two discriminator networks $D_A$ and $D_B$. The generator  $G_{AB}$ is used to translate the given input image from domain $A$ into the output image of domain $B$ and the generator $G_{BA}$ is used to transform an input sample from domain $B$ into the output sample in domain $A$. The discriminator $D_A$ is used to differentiate between the real and the generated image in domain $A$ and in the similar fashion the discriminator $D_B$ is used to differentiate between the real and the generated image in domain $B$. The Real\_Images ($R_A$ and $R_B$) from the domains $A$ and $B$ are given to the generators $G_{AB}$ and $G_{BA}$ to generate the Synthesized\_Images ($Syn_B$ and $Syn_A$) in domains $B$ and $A$, respectively as,

\begin{equation}
\label{syn_imagesB}
Syn_B=G_{AB}(R_A)
\end{equation}
\begin{equation}
\label{syn_imagesA}
Syn_A=G_{BA}(R_B).
\end{equation}

The Synthesized\_Images ($Syn_B$ and $Syn_A$) are given to the generators $G_{BA}$ and $G_{AB}$ to generate the Cycled\_Images ($Cyc_A $ and $Cyc_B)$, respectively as,

\begin{equation}
\label{cycled_imagesA}
Cyc_A=G_{BA}(Syn_B)=G_{BA}((G_{AB}(R_A))
\end{equation}
\begin{equation}
\label{cycled_imagesB}
Cyc_B=G_{AB}(Syn_A)=G_{AB}(G_{BA}(R_B))
\end{equation}
where, $Cyc_A$ is the Cycled\_Image in domain $A$ and $Cyc_B$ is the Cycled\_Image in domain $B$. As shown in the Fig. \ref{fig_CDGAN_architecture}, the overall work of the proposed CDGAN method is to read two Real\_Images ($R_A$ and $R_B$) as input, one from the domain $A$ and another from the domain $B$. These images $R_A$ and $R_B$ are first translated into the Synthesized\_Images ($Syn_B$ and $Syn_A$) of other domains $B$ and $A$ by giving them to the generators $G_{AB}$ and $G_{BA}$, respectively. Later the translated Synthesized\_Images ($Syn_B$ and $Syn_A$) from the domains $B$ and $A$ are again given to the generators $G_{BA}$ and $G_{AB}$ respectively, to get the translated Cycled\_Images ($Cyc_A$ and $Cyc_B$) in domains $A$ and $B$, respectively. 

The proposed CDGAN method is able to translate the input image $R_A$ from domain $A$ into the image $Syn_B$ in another domain $B$, such that the $Syn_B$ has to be look like same as the $R_B$. In the similar fashion the input image $Real\_B$ is translated into the image $Syn_A$, such that it also looks like same as the $R_A$. The difference between the input real images and the translated synthesized images should be minimized in order to get the more realistic generated images. Thus, the suitable loss functions are needed to be used.

\subsection{Objective Functions}
The proposed CDGAN method as shown in the Fig. \ref{fig_CDGAN_architecture} consists of five loss functions, namely Adversarial loss, Synthesized loss,  Cycle-consistency loss, Cyclic-Synthesized loss and Cyclic-Discriminative (CD) Adversarial loss.

\subsubsection{Adversarial Loss}
\label{LSGAN_Loss}
The least-squares loss introduced in LSGAN \cite{LSGAN} is used as an objective in Adversarial loss instead of the negative log  likelihood objective in the vanilla GAN \cite{gan} for more stabilized training. The Adversarial loss is calculated between the fake images generated by the generator network against the decision by the discriminator network either it is able to distinguish it as real or fake. The generator network tries to generate the fake image which looks like same as the real image. The real and the generated images are distinguished using the discriminator network. In the proposed CDGAN, Synthesized\_Image ($Syn_B$) in domain $B$ is generated from the generator $G_{AB}: A \rightarrow B$ by using the Real\_Image ($R_A$) of domain $A$. The Real\_Image ($R_B$) and the Synthesized\_Image ($Syn_B$) in domain $B$ are distinguished by the discriminator $D_B$. It is written as, 

\begin{equation}
\begin{split}
\label{LSGANB}
\mathcal{L}_{LSGAN_B}(G_{AB}, D_B, A, B) &=  \scriptstyle{E}_{B \sim P_{data}(B)} [(D_B(R_B)-1)^2]  +\\ \scriptstyle{E}_{A \sim P_{data}(A)} [D_B(G_{AB}(R_A))^2].
\end{split}
\end{equation}
where, $\mathcal{L}_{LSGAN_B}$ is the Adversarial loss in domain $B$.
In the similar fashion, the Synthesized\_Image ($Syn_A$) in domain $A$ is generated from the Real\_Image ($R_B$) of domain $B$ by using the generator network $G_{BA}: B \rightarrow A$. The Real\_Image ($R_A$) and the Synthesized\_Image ($Syn_A$) in domain $A$ are differentiated by using the discriminator network $D_A$. It is written as,

\begin{equation}
\begin{split}
\label{LSGANA}
\mathcal{L}_{LSGAN_A}(G_{BA}, D_A, B, A) &=  \scriptstyle{E}_{A \sim P_{data}(A)} [(D_A(R_A)-1)^2]
+ \\\scriptstyle{E}_{B \sim P_{data}(B)} [D_A(G_{BA}(R_B))^2].
\end{split}
\end{equation}
where, $\mathcal{L}_{LSGAN_A}$ is the Adversarial loss in domain $A$.
Adversarial loss is used to learn the distributions of the input data during training and to produce the real looking images in testing with the help of that learned distribution. The Adversarial loss tries to eliminate the problem of outputting blurred images, some artifacts are still present.

\subsubsection{Synthesized Loss}
\label{Syn_Loss}
Image-to-image transformation is not only aimed to transform the input image from source domain to target domain, but also to generate the output image as much as close to the original image in the target domain. To fulfill the later one the Synthesized loss is introduced in \cite{ps2-man}. It computes the $L_1$ loss in domain $A$ between the Real\_Image ($R_A$) and the Synthesized\_Image ($Syn_A$) and given as,

\begin{equation}
\label{LSA}
\mathcal{L}_{Syn_A}=\left \| R_A-Syn_A \right \|_1=\left \| R_A-G_{BA}(R_B))\right\|_1
\end{equation}
where, $\mathcal{L}_{Syn_A}$ is the Synthesized loss in domain $A$, $R_A$ and $Syn_A$ are the Real and Synthesized images in domain $A$.

In the similar fashion, the $L_1$ loss in domain $B$ between the Real\_Image ($R_B$) and the Synthesized\_Image ($Syn_B$) is computed as the Synthesized loss and given as, 

\begin{equation}
\label{LSB}
\mathcal{L}_{Syn_B}=\left \| R_B-Syn_B \right \|_1=\left \| R_B-G_{AB}(R_A)\right\|_1
\end{equation}
where, $\mathcal{L}_{Syn_B}$ is the Synthesized loss in domain $B$, $R_B$ and $Syn_B$ are the Real and the Synthesized images in domain $B$. Synthesized loss helps to generate the fake output samples closer to the real samples in the target domain.

\subsubsection{Cycle-consistency Loss}
\label{Cycle-consistency_Loss}
To reduce the discrepancy between the two different domains, the Cycle-consistency loss is introduced in \cite{CyclicGAN}. The $L_1$ loss in domain $A$ between the Real\_Image ($R_A)$ and the Cycled\_Image ($Cyc_A$) is computed as the Cycle-consistency loss and defined as,

\begin{equation}
\label{LCYCA}
\mathcal{L}_{cyc_A}=\left \| R_A-Cyc_A \right \|_1=\left \| R_A-G_{BA}(G_{AB}(R_A))\right\|_1
\end{equation}
where, $\mathcal{L}_{Cyc_A}$ is Cycle-consistency loss in domain $A$, $R_A$ and $Cyc_A$ are the Real and Cycled images in domain $A$. 
In the similar fashion, the $L1$ loss in domain $B$ between the Real\_Image ($R_B$) and the Cycled\_Image ($Cyc_B$) is computed as the Cycle-consistency loss and defined as,
\begin{equation}
\label{Loss_Comparison_table}
\mathcal{L}_{cyc_B}=\left \| R_B-Cyc_B \right \|_1=\left \| R_B-G_{AB}(G_{BA}(R_B))\right\|_1
\end{equation}
where, $\mathcal{L}_{Cyc_B}$ is Cycle-consistency loss in domain $B$, $R_B$ and $Cyc_B$ are the Real and the Cycled images in domain $B$. The Cycle-consistency losses, i.e., $\mathcal{L}_{cyc_A}$ and $\mathcal{L}_{cyc_B}$ used in the objective function act as both forward and backward consistencies. These two Cycle-consistency losses are also included in the objective function of the proposed CDGAN method. The scope of different mapping functions for larger networks is reduced by these losses. They also act as the regularizer for learning the network parameters.

\begin{table*}
\caption{Showing relationship between the six benchmark methods and the proposed CDGAN method in terms of losses. **DualGAN is similar to CycleGAN. The tick mark represents the presence of a loss in a method.}
\label{loss_comparison_table}
\centering
\begin{tabular}{c c c c c c c c c c c c}
\hline
\textbf{Methods} & \multicolumn{8}{ c }{\textbf{Losses}}  \\ 
\cline{2-11}
&  $\mathbf{\mathcal{L}_{LSGAN_A}}$ & $\mathbf{\mathcal{L}_{LSGAN_B}}$ & $\mathbf{\mathcal{L}_{Syn_A}}$ & $\mathbf{\mathcal{L}_{Syn_B}}$ & $\mathbf{\mathcal{L}_{Cyc_A}}$ & $\mathbf{\mathcal{L}_{Cyc_B}}$ & $\mathbf{\mathcal{L}_{CS_A}}$ & $\mathbf{\mathcal{L}_{CS_B}}$ & $\mathbf{\mathcal{L}_{CDGAN_A}}$ & $\mathbf{\mathcal{L}_{CDGAN_B}}$ \\
 \hline
GAN \cite{gan} & $\checkmark$ & $\checkmark$ & &  & & & & & &  \\

Pix2Pix \cite{cGAN} & $\checkmark$ & $\checkmark$ & $\checkmark$ & $\checkmark$ & & & & & &  \\

**DualGAN \cite{dualGAN} & $\checkmark$ & $\checkmark$ &  & & $\checkmark$ & $\checkmark$& && &  \\

CycleGAN \cite{CyclicGAN}& $\checkmark$ & $\checkmark$ &  & & $\checkmark$ & $\checkmark$& && &  \\

PS2GAN \cite{ps2-man}& $\checkmark$ & $\checkmark$ & $\checkmark$ & $\checkmark$ & $\checkmark$& $\checkmark$& && & \\

CSGAN \cite{csgan}& $\checkmark$ &$\checkmark$&& &$\checkmark$ & $\checkmark$&$\checkmark$ &$\checkmark$ & &  \\
\hline
CDGAN (Ours) & $\checkmark$& $\checkmark$ & $\checkmark$ & $\checkmark$&$\checkmark$ & $\checkmark$&$\checkmark$ & $\checkmark$&$\checkmark$ & $\checkmark$ 
\\
\hline
\end{tabular}
\end{table*}

\subsubsection{Cyclic-Synthesized Loss}
\label{CS_loss}
The Cyclic-Synthesized loss introduced in CSGAN \cite{csgan}, which is computed as the $L_1$ loss between the Synthesized\_Image ($Syn_A$) and the Cycled\_Image ($Cyc_A$) in domain $A$ and defined as,
\begin{equation}
\begin{split}
\label{LCSA}
\mathcal{L}_{CS_A}=\left \| Syn_A-Cyc_A \right \|_1= \\ \left \| G_{BA}(R_B)-G_{BA}(G_{AB}(R_A))\right\|_1
\end{split}
\end{equation}
where, $\mathcal{L}_{CS_A}$ is the Cyclic-Synthesized loss, $Syn_A$ and $Cyc_A$ are the Synthesized and Cycled images in domain $A$. 

Similarly, the $L1$ loss in domain $B$ between the Synthesized\_Image ($Syn_B$) and the Cycled\_Image ($Cyc_B$) is computed as the Cyclic-Synthesized loss and defined as,

\begin{equation}
\label{LCSB}
\begin{split}
\mathcal{L}_{CS_B}=\left \| Syn_B-Cyc_B \right \|_1= \\ \left \| G_{AB}(R_A)-G_{AB}(G_{BA}(R_B))\right\|_1
\end{split}
\end{equation}
where, $\mathcal{L}_{CS_B}$ is the Cyclic-Synthesized loss, $Syn_B$ and $Cyc_B$ are the Real and the Cycled images in domain $B$. These two Cyclic-Synthesized losses are also included in the objective function of the proposed CDGAN method.  Although, all the above mentioned losses are included in the objective function of the proposed CDGAN method, still there is a lot of scope to improve the quality of the images generated and to remove the unwanted artifacts produced in the resulting images. To fulfill that scope, a new loss called Cyclic-Discriminative Adversarial loss is proposed in this paper to generate the high quality images with reduced artifacts.

\subsubsection{Proposed Cyclic-Discriminative Adversarial Loss}
\label{CDGAN_Loss}
The Cyclic-Discriminative Adversarial loss proposed in this paper is calculated as the adversarial loss between the Real\_Images ($R_A$ and $R_B$) and the Cycled\_Images ($Cyc_A$ and $Cyc_B$). 
The adversarial loss in domain $A$ between the Cycled\_Image ($Cyc_A$) and the Real\_Image ($R_A$) by using the generator $G_{BA}$ and the discriminator $D_A$ is computed as the Cyclic-Discriminative adversarial loss in domain $A$ and defined as,
\begin{equation}
\begin{split}
\label{CDGANA}
\mathcal{L}_{CDGAN_A}(G_{BA}, D_A, B, A) &=  \scriptstyle{E}_{A \sim P_{data}(A)} [(D_A(R_A)-1)^2]  
\\+\scriptstyle{E}_{B \sim P_{data}(B)} [D_A(G_{BA}(Syn_B))^2].
\end{split}
\end{equation}
where, $\mathcal{L}_{CDGAN_A}$ is the Cyclic-Discriminative adversarial loss in domain $A$, $Cyc_A$ (=$G_{BA}(Syn_B)$) and $R_A$ are the cycled and real image respectively.

Similarly, the adversarial in domain $B$ loss between the Cycled\_Image ($Cyc_B$) and the Real\_Image ($R_B$) by using the generator $G_{AB}$ and the discriminator $D_B$ is computed as the Cyclic-Discriminative adversarial loss in domain $B$ and defined as,
\begin{equation}
\begin{split}
\label{CDGANB}
\mathcal{L}_{CDGAN_B}(G_{AB}, D_B, A, B) &=  \scriptstyle{E}_{B \sim P_{data}(B)} [(D_B(R_B)-1)^2] \\ +\scriptstyle{E}_{A \sim P_{data}(A)} [D_B(G_{AB}(Syn_A))^2].
\end{split}
\end{equation}
where, $\mathcal{L}_{CDGAN_B}$ is the Cyclic-Discriminative adversarial loss in domain $B$, $Cyc_B$ (=$G_{AB}(Syn_A)$) and $R_B$ are the cycled and real image respectively.
Finally, we combine all the losses to have the CDGAN Objective function. The relationship between the proposed CDGAN method and six benchmark methods are shown in Table \ref{loss_comparison_table} for better understanding.

\subsection{CDGAN Objective Function}
\label{Final_Loss}
The CDGAN method final objective function combines existing Adversarial loss, Synthesized loss, Cycle-consistency loss and Cyclic-Synthesized loss along with the proposed Cyclic-Discriminative Adversarial loss  as follows,
\begin{equation}
\label{LFINAL}
\begin{split}
\mathcal{L}_(G_{AB}, G_{BA}, D_A,
D_B)=\mathcal{L}_{LSGAN_A}+\mathcal{L}_{LSGAN_B} \\ +\mu_A\mathcal{L}_{Syn_A} +\mu_B{\mathcal{L}_{Syn_B}}+\lambda_A\mathcal{L}_{cyc_A}  +\lambda_B{\mathcal{L}_{cyc_B}} \\+\omega_A\mathcal{L}_{CS_A}+\omega_B\mathcal{L}_{CS_B} +\mathcal{L}_{CDGAN_A} +{\mathcal{L}_{CDGAN_B}}.
\end{split}
\end{equation}
\label{LCYCA1}
where $\mathcal{L}_{CDGAN_A}$ and $\mathcal{L}_{CDGAN_B}$ are the proposed Cyclic-Discriminative Adversarial losses described in subsection \ref{CDGAN_Loss}; $\mathcal{L}_{LSGAN_A}$ and $\mathcal{L}_{LSGAN_B}$ are the adversarial losses,  $\mathcal{L}_{Syn_A}$ and $\mathcal{L}_{Syn_B}$ are the Synthesized losses, $\mathcal{L}_{cyc_A}$ and $\mathcal{L}_{cyc_B}$ are the Cycle-consistency losses and $\mathcal{L}_{CS_A}$ and $\mathcal{L}_{CS_B}$ are the Cyclic-Synthesized losses explained in the subsections \ref{LSGAN_Loss}, \ref{Syn_Loss}, \ref{Cycle-consistency_Loss} and \ref{CS_loss}, respectively. The $\mu_A$, $\mu_B$, $\alpha_A$, $\alpha_B$, $\omega_A$ and $\omega_B$ are the weights for the different losses. The values of these weights are set empirically.
  
\section{Experimental Setup} \label{Exp Setup}
This section is devoted to describe the datasets used in the experiment with train and test partitions, evaluation metrics used to judge the performance and the training settings used to train the models. This section also describes the network architectures and baseline GAN models.

\subsection{DataSets} \label{DataSets}
For the experimentation we used the following three different baseline datasets meant for the image-to-image transformation task.
 
\subsubsection{CUHK Face Sketch Dataset} The CUHK\footnote{http://mmlab.ie.cuhk.edu.hk/archive/facesketch.html}dataset consists of $188$ students face photo-sketch image pairs. The cropped version of the data with the dimension $250 \times 200$ is used in this paper. Out of the total $188$ images of the dataset, $100$ images are used for the training and the remaining images are used for the testing.

\subsubsection{CMP Facades Dataset} The CMP Facades\footnote{http://cmp.felk.cvut.cz/~tylecr1/facade/} dataset contains a total of $606$ labels and corresponding facades image pairs. Out of the total $606$ images of this dataset, $400$ images are used for the training and the remaining images are used for the testing.


\subsubsection{RGB-NIR Scene Dataset} 
The RGB-NIR\footnote{https://ivrl.epfl.ch/research-2/research-downloads/supplementary\_material-cvpr11-index-html/} dataset consists of $477$ images taken from $9$ different categories captured in both RGB and Near-infrared (NIR) domains. Out the total $477$ images of this dataset, $387$ images are used for the training and the remaining $90$ images are used for the testing.

\begin{figure*}[t]
\begin{center}
\includegraphics[width=16cm, height=8.5cm]{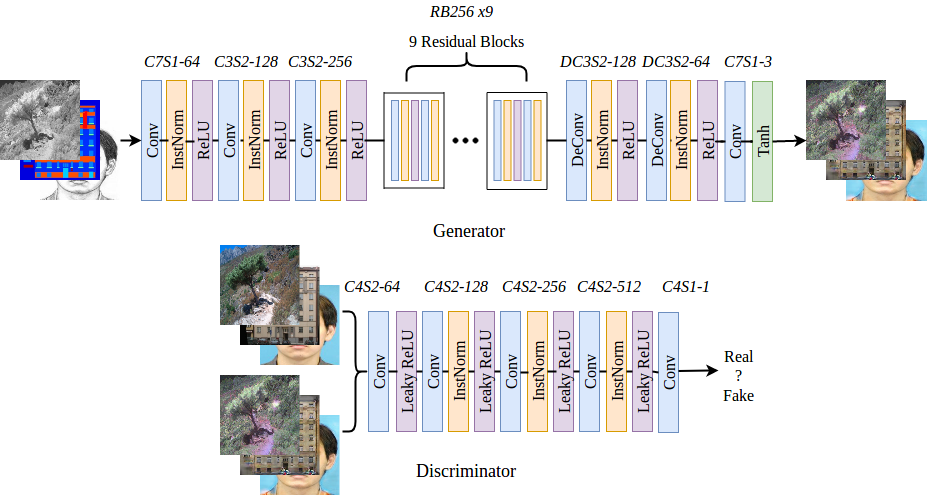}
\caption{Generator and Discriminator Network architectures. }
\label{fig_Gen-Disc}
\end{center}
\end{figure*}

\subsection{Training Information}
The network is trained with the input images of fixed size $256\times256$, each image is resized from the arbitrary size of the dataset to the fixed size of $256\times256$. The network is initialized with the same setup in \cite{cGAN}. Both the generator and the discriminator networks are  trained with batch size $1$ from scratch to $200$ epochs. Initially, the learning rate is set to $0.0002$ for the first $100$ epochs and linearly decaying down it to $0$ over the next $100$ epochs. The weights of the network are initialized with the Gaussian distribution having mean $0$ and standard deviation $0.02$. The network is optimized with the Adam solver \cite{adam} having the momentum term $\beta_1$ as $0.5$ instead of $0.9$, because as per \cite{dcgan} the momentum term $\beta_1$ as $0.9$ or higher values can cause to substandard network stabilization for the image-to-image transformation task. The values of the weight factors for the proposed CDGAN method,  $\mu_A$ and $\mu_B$ are set to $15$, $\lambda_A$ and $\lambda_B$ are set to $10$ and $\omega_A$ and $\omega_B$ are set to $30$ (see Equation \ref{LFINAL}). For the comparison methods, the values for the weight factors are taken from their source papers.

\subsection{Network Architectures}
Network architectures of the generator and the discriminator as shown in Fig.\ref{fig_Gen-Disc} and used in this paper are taken from \cite{CyclicGAN}. Following is the $9$ residual blocks used in the generator network: $C7S1\_64$, $C3S2\_128$, $C3S2\_256$, $RB256\times9$, $DC3S2\_128$, $DC3S2\_64$, $C7S1\_3$, where, $C7S1\_f$ represents a $7\times7$ Convolutional layer with $f$ filters and stride $1$, $C3S2\_f$ represents a $3\times3$ Convolutional\_InstanceNorm\_ReLU layer with $f$ filters and stride $2$, $RBf\times n$ represents $n$ residual blocks consist of two Convolutional layers with $f$ filters for both layers, and $DC3S2\_f$ represents $3\times3$ DeConvolution\_InstanceNorm\_ReLU layer with $f$ filters and stride $\frac{1}{2}$.

For the discriminator, we use $70\times70$ PatchGAN from \cite{cGAN}. The discriminator network consists of: $C4S2\_64$, $C4S2\_128$, $C4S2\_256$, $C4S2\_512$, $C4S1\_1$, where $C4S2\_f$ represents a $4\times4$ Convolution\_InstanceNorm\_LeakyReLU layer with $f$ filters and stride $2$, and $C4S1-1$ represents a $4\times4$ Convolutional layer with $f$ filters and stride $1$ to produce the final one-dimensional output. The activation function used in this work is the Leaky ReLU with 0.2 slope. The first Convolution layer does not include the InstanceNorm.
\label{LCYCA3}

\subsection{Evaluation Metrics}
To better understand the improved performance of the proposed CDGAN method, both the quantitative and qualitative metrics are used in this paper. Most widely used image quality assessment metrics for image-to-image transformation like Peak Signal to Noise Ratio (PSNR), Mean Square Error (MSE), and Structural Similarity Index (SSIM)  \cite{SSIM} are used under quantitative evaluation. Learned Perceptual Image Patch Similarity (LPIPS) proposed in \cite{LPIPS} is also used for calculating the perceptual similarity. The distance between the ground truth and the generated fake image is computed as the LPIPS score. The enhanced quality images generated by the proposed CDGAN method along with the six benchmark methods and ground truth are also compared in the result section.
\begin{table*}[t]
\caption{The quantitative comparison of the results of the proposed CDGAN with different state-of-the art methods trained on CUHK, FACADES and RGB-NIR Scene Datasets. The average scores for the SSIM, MSE, PSNR and LPIPS metrics are reported. Best results are highlighted in bold and second best results are shown in italic font.}
\label{metrics_comparison_table}
\begin{center}
\scalebox{1.07}{
\begin{tabular}{c c c c c c c c c }
\hline
\\
\textbf{Datasets} & \textbf{Metrics} & \multicolumn{7}{ c }{\textbf{Methods}}  \\ 
\cline{3-9}
\\
& &GAN \cite{gan} & Pix2Pix \cite{cGAN} & DualGAN \cite{dualGAN} & CycleGAN \cite{CyclicGAN} &PS2GAN \cite{ps2-man} & CSGAN \cite{csgan}& CDGAN 
\\  \\
 \hline
 \\
\multirow{4}{*}{CUHK} 

& SSIM  & $0.5398$ & $0.6056$ &$0.6359$ &$0.6537$  &$0.6409$ &$\mathit{0.6616}$ &$\mathbf{0.6852}$  \\

&MSE  & $94.8815$ & $89.9954$ &$85.5418$ & $89.6019$ &$86.7004$ &$\mathit{84.7971}$ &$\mathbf{82.9547}$  \\

&PSNR & $28.3628$ & $28.5989$ &$28.8351$  &$28.6351$ &$28.7779$ & $\mathit{28.8693}$&$\mathbf{28.9801}$   \\

&LPIPS & $0.157$ & $0.154$ & $0.132$ &$0.099$ & $0.098$ & $\mathit{0.094}$& $\mathbf{0.090}$ \\ \\
\hline

\\

\multirow{4}{*}{FACADES} 

& SSIM  & $0.1378$ & $0.2106$ &$0.0324$ &$0.0678$  &$0.1764$ &$\mathit{0.2183}$ &$\mathbf{0.2512}$  \\

&MSE  & $103.8049$ & $\mathit{101.9864}$ &$105.0175$ & $104.3104$ &$102.4183$ &$103.7751$ &$\mathbf{101.5533}$  \\

&PSNR & $27.9706$ & $\mathit{28.0569}$ &$27.9187$  &$27.9849$ &$28.032$ & $27.9715$&$\mathbf{28.0761}$   \\

&LPIPS & $0.252$ & $\mathit{0.216}$ & $0.259$ &$0.248$ & $0.221$ & $0.22$& $\mathbf{0.215}$  \\ \\
\hline
\\

\multirow{4}{*}{RGB-NIR} 
&SSIM  & $0.4788$ & $0.575$ &$-0.0126$ &$0.5958$  &$\mathit{0.597}$ &$0.5825$ &$\mathbf{0.6265}$  \\

&MSE  & $101.6426$ & $100.0377$ &$105.4514$ & $98.2278$ &$\mathit{97.5769}$ &$98.704$ &$\mathbf{96.5412}$  \\

&PSNR & $28.072$ & $28.1464$ &$27.9019$  &$28.2574$ &$28.2692$ & $\mathit{28.2159}$&$\mathbf{28.3083}$   \\

&LPIPS & $0.243$ & $0.182$ & $0.295$ &$0.18$ & $\mathit{0.166}$ & 0.178& $\mathbf{0.147}$  \\  \\
\hline

\end{tabular}
}
\end{center}
\end{table*}

\subsection{Baseline Methods}
We compare the proposed CDGAN model with six benchmark models, namely, GAN\cite{gan}, Pix2Pix\cite{cGAN}, DualGAN\cite{dualGAN}, CycleGAN\cite{CyclicGAN}, PS2MAN \cite{ps2-man} and CSGAN \cite{csgan} to demonstrate its significance. All the above mentioned comparisons are made in paired setting only. 
 
\subsubsection{GAN}
The original vanilla GAN proposed in \cite{gan} is used to generate the new samples from the learned distribution function with the given noise vector. Whereas, the GAN used for comparison in this paper is implemented for image-to-image translation from the Pix2Pix\footnote{https://github.com/phillipi/pix2pix \label{Pix2Pix}}\cite{cGAN} by removing the $L_1$ loss and keeping only the adversarial loss.
\subsubsection{Pix2Pix}
For this method, the code provided by the authors in Pix2Pix\cite{cGAN} is used for generating the result images with the same default settings.
\subsubsection{DualGAN}
For this method, the code provided by the authors in DualGAN\footnote{https://github.com/duxingren14/DualGAN}\cite{dualGAN} is used for generating the result images with the same default settings from the original code.
\subsubsection{CycleGAN}
For this method, the code provided by the authors in CycleGAN\footnote{https://github.com/junyanz/pytorch-CycleGAN-and-pix2pix}\cite{CyclicGAN} is used for generating the result images with the same default settings.
\subsubsection{PS2GAN}
In this method, the code is implemented by adding the synthesized loss to the existing losses of CycleGAN\cite{CyclicGAN} method. For a fair comparison of the state-of-the-art methods with proposed CDGAN method, the PS2MAN \cite{ps2-man} method originally proposed with multiple adversarial networks is modified to single adversarial networks, i.e., PS2GAN.
\subsubsection{CSGAN}
For this method, the code provided by the authors in CSGAN\footnote{https://github.com/KishanKancharagunta/CSGAN}\cite{csgan} is used for generating the result images with the same default settings.

\begin{figure*}[t]
\begin{center}
\includegraphics[width=18.0 cm, height=8.0 cm]{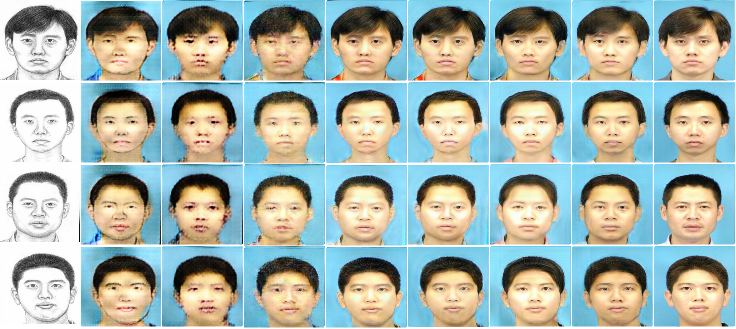}
\caption{The qualitative comparison of generated faces for sketch-to-photo synthesis over CUHK Face Sketch dataset. The Input, GAN, Pix2Pix, DualGAN, CycleGAN, PS2GAN, CSGAN, CDGAN and Ground Truth images are shown from left to right, respectively. The CDGAN generated faces have minimal artifacts and look like more realistic with sharper images.}
\label{cuhk_fig}
\end{center}
\end{figure*}

\begin{figure*}[!t]
\begin{center}
\includegraphics[width=18.0 cm, height=8.0 cm]{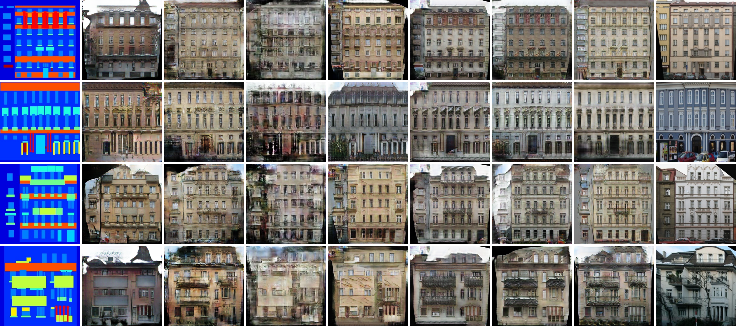}
\caption{The qualitative comparison of generated building images for label-to-building transformation over FACADES dataset. The Input, GAN, Pix2Pix, DualGAN, CycleGAN, PS2GAN, CSGAN, CDGAN and Ground Truth images are shown from left to right, respectively. The CDGAN generated building images have minimal artifacts and looking more realistic with sharper images.}
\label{facades_fig}
\end{center}
\end{figure*}

\begin{figure*}[!t]
\begin{center}
\includegraphics[width=18.0 cm, height=8.0 cm]{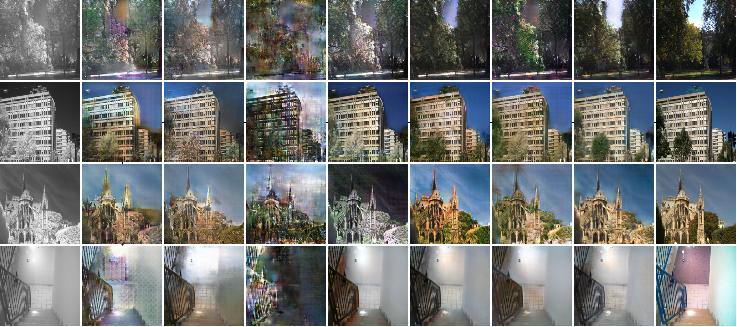}
\caption{The qualitative comparison of generated RGB scenes for RGB-to-NIR scene image transformation over RGB-NIR scene dataset.  The Input, GAN, Pix2Pix, DualGAN, CycleGAN, PS2GAN, CSGAN, CDGAN and Ground Truth images are shown from left to right, respectively. The CDGAN generated RGB scenes have minimal artifacts and looking more realistic with sharper images.}
\label{rgb-nir-scene_fig}
\end{center}
\end{figure*}

\begin{table*}[t]
\caption{The quantitative results comparison between the proposed CDGAN method and different state-of-the art methods trained on CUHK, FACADES and RGB-NIR Scene Datasets.
The average scores for the SSIM, MSE, PSNR and LPIPS scores are reported. The '+' symbol represents the presence of Cyclic-Discriminative Adversarial loss. The results with Cyclic-Discriminative Adversarial loss are highlighted in italic and best results produced by the proposed CDGAN method are highlighted in bold.}
\label{loss_comparison_table_ablation}
\centering
\begin{tabular}{c c c c c c c c c c c }
\hline
\\
\textbf{Datasets} & \textbf{Metrics} & \multicolumn{9}{ c }{\textbf{Methods}}\\
\cline{3-11}
\\
& & DualGAN  &  DualGAN+ & CycleGAN  & CycleGAN+ &PS2GAN  &PS2GAN+ & CSGAN & CSGAN+&Our Method \\ \\
\hline
\\
\multirow{4}{*}{CUHK} & SSIM  &  $0.6359$ & $0.6357$ &$0.6537$ & $0.6309$ & $0.6409$ &$0.6835$  & $0.6616$ & $0.6762$ & $\mathbf{0.6852}$  \\
& MSE  &  $85.5418$ & $88.1905$ & $89.6019$ & $87.2625$ &$86.7004$ & $83.6166$ & $84.7971$ & $83.7534$ & $\mathbf{82.9547}$  \\
& PSNR & $28.8351$  & $28.6351$ & $28.7444$ &$28.6932$ &$28.7779$ & $28.9465$& $28.8693$ & $28.947$ & $\mathbf{28.9801}$   \\
& LPIPS &  $0.132$ & $0.104$ &$0.099$ & $0.109$& $0.098$ &$0.106$ & $0.094$ & $0.105$& $\mathbf{0.090}$ \\ \\
\hline
\\
\multirow{4}{*}{FACADES} & SSIM  &  $0.0324$ & $0.0542$ &$0.0678$ & $0.0249$ & $0.1764$ &$0.0859$  & $0.2183$ & $0.2408$ & $\mathbf{0.2512}$  \\
& MSE  &  $105.0175$ & $105.0602$ &$104.3104$ & $105.0355$ & $102.4183$ & $104.4056$ & $103.775$ & $101.4973$ & $\mathbf{101.5533}$  \\
& PSNR & $27.9187$ & $27.9172$  & $27.9489$ & $27.9185$ &$28.032$ &$27.9446$ & $27.9715$& $28.0767$  & $\mathbf{28.0761}$   \\
& LPIPS &  $0.259$ & $0.251$ & $0.248$ & $0.264$&$0.221$ &$0.234$ & $0.22$ & $0.214$& $\mathbf{0.215}$ \\ \\
\hline
\\
\multirow{4}{*}{RGB-NIR} & SSIM  &  $-0.0126$ & $0.4088$ &$0.5958$ & $0.5795$ & $0.597$ &$0.6154$  & $0.5825$ & $0.6129$ & $\mathbf{0.6265}$  \\
& MSE  &  $105.4514$ & $102.8754$ &$98.2278$ & $98.1801$ & $97.5769$ & $97.3519$ & $98.704$ & $96.8126$ & $\mathbf{96.5412}$ \\
& PSNR & $27.9109$  & $28.0116$ &$28.2574$ & $28.2619$ &$28.2692$ & $28.2874$& $28.2159$ & $28.2933$ & $\mathbf{28.3083}$   \\
& LPIPS &  $0.295$ & $0.221$ &$0.18$ & $0.189$& $0.166$ &$0.174$ & $0.178$ & $0.152$& $\mathbf{0.147}$ \\ \\
\hline
\end{tabular}
\end{table*}

\section{Result Analysis}
\label{Result Analysis}
This section is dedicated to analyze the results produced by the introduced CDGAN method. To better express the improved performance of the proposed CDGAN method, we consider both the quantitative and qualitative evaluation. The proposed CDGAN method is compared against the six benchmark methods, namely, GAN, Pix2Pix, DualGAN, CycleGAN, PS2GAN and CSGAN. We also perform an ablation study over the proposed cyclic-discriminative adversarial loss with DualGAN, CycleGAN, PS2GAN and CSGAN to investigate its suitability with existing losses.

\subsection{Quantitative Evaluation}
For the quantitative evaluation of the results, four baseline quantitative measures like SSIM, MSE, PSNR and LPIPS are used. The average scores of these four metrics are calculated for all the above mentioned methods. The results over CUHK, FACADES and RGB-NIR scene datasets are shown in Table \ref{metrics_comparison_table}. The larger SSIM and PSNR scores and the smaller MSE and LPIPS scores indicate the generated images with better quality. The followings are the observations from the results of this experiment:

\begin{itemize}
  \item The proposed CDGAN method over CUHK dataset achieves an improvement of $26.93\%$, $13.14\%$, $7.75\%$, $4.81\%$, $6.91\%$ and $3.56\%$ in terms of the SSIM metric and reduces the MSE at the rate of $12.57\%$, $7.82\%$, $3.02\%$, $7.41\%$, $4.32\%$ and $2.17\%$ as compared to GAN, Pix2Pix, DualGAN, CycleGAN, PS2GAN and CDGAN, respectively.
  \item For the FACADES dataset, the proposed CDGAN method achieves an improvement of $82.29\%$, $19.27\%$, $675.30\%$, $270.50\%$, $42.40\%$ and $15.07\%$ in terms of the SSIM metric and reduces the MSE at the rate of $2.16\%$, $0.42\%$, $3.29\%$, $2.64\%$, $0.84\%$ and $2.14\%$ as compared to GAN, Pix2Pix, DualGAN, CycleGAN, PS2GAN and CDGAN, respectively.
  \item The proposed CDGAN method exhibits an improvement of $30.84\%$, $8.95\%$, $5072.22\%$, $5.15\%$, $4.94\%$ and $7.55\%$ in the SSIM score, whereas shows the reduction in the MSE score by $5.01\%$, $3.49\%$, $8.44\%$, $1.71\%$, $1.06\%$ and $2.19\%$ as compared to GAN, Pix2Pix, DualGAN, CycleGAN, PS2GAN and CDGAN, respectively, over RGB-NIR scene dataset. 
  \item The performance of DualGAN is very bad over FACADES and RGB-NIR scene datasets because, these tasks involve high semantics-based labeling. The similar behavior of DualGAN is also observed by its original authors \cite{dualGAN}. 
  \item The PSNR and LPIPS measures over CUHK, FACADES and RGB-NIR scene datasets also show the reasonable improvement due to the proposed CDGAN method compared to other methods. 
\end{itemize} 

From the above mentioned comparisons over three different datasets, it is clearly understandable that the proposed CDGAN method generates more structurally similar, less pixels to pixel noise and perceptually real looking images with reduced artifacts as compared to the state-of-the-art approaches. Note that the SSIM of DualGAN on RGB-NIR dataset is negative as SSIM uses the co-variance between two images which can be negative, if the similarity in images is low.

\subsection{Qualitative Evaluation}
In order to show the improved quality of the output images produced by the proposed CDGAN, we compare and show few sample image results generated by the CDGAN against six state-of-the art methods. These comparisons over CUHK, FACADES and RGB-NIR scene datasets are shown in Fig. \ref{cuhk_fig}, \ref{facades_fig} and \ref{rgb-nir-scene_fig}, respectively. The followings are the observations and analysis drawn from these qualitative results:
\begin{itemize}
  \item From Fig. \ref{cuhk_fig}, it can be observed that the images generated by the CycleGAN, PS2GAN and CSGAN methods, contain the reflections on the faces for the sample faces of CUHK dataset. Whereas, the proposed CDGAN method is able to eliminate this effect due to its increased discriminative capacity. Moreover, the facial attributes in generated faces such as eyes, hair and face structure, are generated without the artifacts using the proposed CDGAN method.
  \item The proposed CDGAN method generates the building images over FACADES dataset with enhanced textural detail information such as window and door sizes and shapes. As shown in Fig. \ref{facades_fig}, the buildings generated by the proposed CDGAN method consist more structural information. In the first row of Fig. \ref{facades_fig}, the building generated by the CDGAN method from the top left corner contains more window information compared to the remaining methods.
  \item The qualitative results over RGB-NIR scene dataset are illustrated in Fig \ref{rgb-nir-scene_fig}). The generated RGB images using the proposed CDGAN method contain more semantic, depth aware and structure aware information as compared to the state-of-the-art methods. It can be seen in the first row of Fig. \ref{rgb-nir-scene_fig} that the proposed CDGAN method is able to generate the grass in green color at the bottom portion of the generated RGB image, where other methods fail. Similarly, it can be also seen in $2^{nd}$ row of Fig. \ref{rgb-nir-scene_fig} that the proposed CDGAN method generates the tree image in front of the building, whereas the compared methods fail to generate. The possible reason for such improved performance is due to the discriminative ability of the proposed method for varying depths of the scene points. It confirms the structure and depth sensitivity of CDGAN method. 
  \item As expected, DualGAN completely fails to produce high semantics labeling based image-to-image transformation tasks over FACADES (see Column $4$ of Fig. \ref{facades_fig}) and RGB-NIR (see Column $4$ of Fig. \ref{rgb-nir-scene_fig}) scene datasets.
\end{itemize}

It is evident from the above qualitative results that the images generated by CDGAN method are more realistic and sharp with reduced artifacts as compared to the state-of-the-art GAN models of image-to-image transformation. 

\subsection{Ablation Study}
In order to analyze the importance of the proposed Cyclic-Discriminative adversarial loss, We conduct an ablation study on losses. We also investigate its dependency on other loss functions such as Adversarial, Synthesized, Cycle-consistency and Cyclic-Synthesized losses. Basically, the Cyclic-Discriminative adversarial loss is added to DualGAN, CycleGAN, PS2GAN, and CSGAN and compared with original results of these methods. The comparison results over the CUHK, FACADES and RGB-NIR scene datasets are shown in Table \ref{loss_comparison_table_ablation}. We observe the following points: 

\begin{itemize}
    \item The proposed Cyclic-Discriminative adversarial loss when added to the DualGAN and CycleGAN have a mix of positive and negative impacts on the generated images as shown in the DualGAN+ and CycleGAN+ columns in the Table \ref{loss_comparison_table_ablation}. Note that the proposed Cyclic-Discriminative adversarial loss is computed between the Real\_Image and the Cycled\_Image. Because the DualGAN+ and CycleGAN+ does not use the synthesized losses, our cyclic-discriminator loss becomes more powerful in these frameworks. It may lead to situation where the generator is unable to fool the discriminator. Thus, the generator may stop learning after a while.
    \item It is also observed that the proposed Cyclic-Discrminative adversarial loss is well suited with the PS2GAN and CSGAN methods as depicted in the Table \ref{loss_comparison_table_ablation}. The improved performance is due to a very tough mini-max game between the powerful generator equipped with synthesized losses and powerful proposed discriminator. Due to this very competitive adversarial learning, the generator is able to produce very high quality images.
    \item It can be seen in the last column of Table \ref{loss_comparison_table_ablation} that the CDGAN still outperforms all other combinations of losses. It confirms the importance of the proposed Cyclic-Discriminative adversarial loss in conjunction with the existing loss functions such as Adversarial, Synthesized, Cycle-consistency and Cyclic-Synthesized losses.
\end{itemize}

This ablation study reveals that the proposed CDGAN method is able to achieve the improved performance when proposed Cyclic-Discriminative adversarial loss is used with synthesized losses.

\section{Conclusion}
\label{Conclusion}
In this paper an improved image-to-image transformation method called CDGAN is proposed. A new Cyclic-Discriminative adversarial loss is introduced to increase the adversarial learning complexity. The introduced Cyclic-Discriminative adversarial loss along with the existing losses are used in CDGAN method. Three different datasets namely CUHK, FACADES and RGB-NIR scene are used for the image-to-image transformation experiments. The experimental quantitative and qualitative results are compared against GAN models including GAN, Pix2Pix, DualGAN, CycleGAN, PS2GAN and CSGAN. It is observed that the proposed method outperforms all the compared methods over all datasets in terms of the different evaluation metrics such as SSIM, MSE, PSNR and LPIPS. The qualitative results also point out the improved generated images in terms of the more realistic, structure preserving, and reduced artifacts. It is also noticed that the proposed method deals better with the varying depths of the scene points. The ablation study over different losses reveals that the proposed loss is better suited with the synthesized losses as it increases the competitiveness between generator and discriminator to learn more semantic features. It is also noticed that the best performance is gained after combining all losses, including Adversarial loss, Synthesized loss, Cycle-consistency loss, Cyclic-Synthesized loss and Cyclic-Discriminative adversarial loss.


\section*{Acknowledgement}
We are grateful to NVIDIA Corporation for donating us the NVIDIA GeForce Titan X Pascal 12GB GPU which is used for this research.

\bibliographystyle{IEEEtran}  
\bibliography{egbib}

\begin{thebibliography}{10}
\providecommand{\url}[1]{#1}
\csname url@samestyle\endcsname
\providecommand{\newblock}{\relax}
\providecommand{\bibinfo}[2]{#2}
\providecommand{\BIBentrySTDinterwordspacing}{\spaceskip=0pt\relax}
\providecommand{\BIBentryALTinterwordstretchfactor}{4}
\providecommand{\BIBentryALTinterwordspacing}{\spaceskip=\fontdimen2\font plus
\BIBentryALTinterwordstretchfactor\fontdimen3\font minus
  \fontdimen4\font\relax}
\providecommand{\BIBforeignlanguage}[2]{{%
\expandafter\ifx\csname l@#1\endcsname\relax
\typeout{** WARNING: IEEEtran.bst: No hyphenation pattern has been}%
\typeout{** loaded for the language `#1'. Using the pattern for}%
\typeout{** the default language instead.}%
\else
\language=\csname l@#1\endcsname
\fi
#2}}
\providecommand{\BIBdecl}{\relax}
\BIBdecl

\bibitem{FSSMAL}
S.~Zhang, R.~Ji, J.~Hu, X.~Lu, and X.~Li, ``Face sketch synthesis by
  multidomain adversarial learning,'' \emph{IEEE transactions on neural
  networks and learning systems}, vol.~30, no.~5, pp. 1419--1428, 2018.

\bibitem{DCFFPSS}
M.~Zhu, J.~Li, N.~Wang, and X.~Gao, ``A deep collaborative framework for face
  photo--sketch synthesis,'' \emph{IEEE transactions on neural networks and
  learning systems}, vol.~30, no.~10, pp. 3096--3108, 2019.

\bibitem{MRBFSPS}
C.~Peng, X.~Gao, N.~Wang, D.~Tao, X.~Li, and J.~Li, ``Multiple
  representations-based face sketch--photo synthesis,'' \emph{IEEE transactions
  on neural networks and learning systems}, vol.~27, no.~11, pp. 2201--2215,
  2015.

\bibitem{DC}
Z.~Cheng, Q.~Yang, and B.~Sheng, ``Deep colorization,'' in \emph{IEEE
  International Conference on Computer Vision}, 2015.

\bibitem{CIC}
R.~Zhang, P.~Isola, and A.~A. Efros, ``Colorful image colorization,'' in
  \emph{European Conference on Computer Vision}, 2016.

\bibitem{HRIPMSNPS}
C.~Yang, X.~Lu, Z.~Lin, E.~Shechtman, O.~Wang, and H.~Li, ``High-resolution
  image inpainting using multi-scale neural patch synthesis,'' in \emph{IEEE
  Conference on Computer Vision and Pattern Recognition}, 2017.

\bibitem{FLIP}
D.~Pathak, P.~Krahenbuhl, J.~Donahue, T.~Darrell, and A.~A. Efros, ``Context
  encoders: Feature learning by inpainting,'' in \emph{IEEE Conference on
  Computer Vision and Pattern Recognition}, 2016, pp. 2536--2544.

\bibitem{PLRTSTSR}
J.~Johnson, A.~Alahi, and L.~Fei-Fei, ``Perceptual losses for real-time style
  transfer and super-resolution,'' in \emph{European Conference on Computer
  Vision}, 2016, pp. 694--711.

\bibitem{VSR}
A.~{Lucas}, S.~{Lopez-Tapiad}, R.~{Molinae}, and A.~K. {Katsaggelos},
  ``Generative adversarial networks and perceptual losses for video
  super-resolution,'' \emph{IEEE Transactions on Image Processing}, pp. 1--1,
  2019.

\bibitem{DLISR}
T.~Guo, H.~S. Mousavi, and V.~Monga, ``Deep learning based image
  super-resolution with coupled backpropagation,'' in \emph{IEEE Global
  Conference on Signal and Information Processing}, 2016, pp. 237--241.

\bibitem{SISRDL}
J.~Chen, X.~He, H.~Chen, Q.~Teng, and L.~Qing, ``Single image super-resolution
  based on deep learning and gradient transformation,'' in \emph{IEEE
  International Conference on Signal Processing}, 2016, pp. 663--667.

\bibitem{ANLAID}
A.~Buades, B.~Coll, and J.-M. Morel, ``A non-local algorithm for image
  denoising,'' in \emph{IEEE Conference on Computer Vision and Pattern
  Recognition}, 2005, pp. 60--65.

\bibitem{RRNRain}
J.~{Liu}, W.~{Yang}, S.~{Yang}, and Z.~{Guo}, ``D3r-net: Dynamic routing
  residue recurrent network for video rain removal,'' \emph{IEEE Transactions
  on Image Processing}, vol.~28, no.~2, pp. 699--712, Feb 2019.

\bibitem{idrcgan}
H.~Zhang, V.~Sindagi, and V.~M. Patel, ``Image de-raining using a conditional
  generative adversarial network,'' \emph{IEEE transactions on circuits and
  systems for video technology}, 2019.

\bibitem{vhrgan}
Y.~Pang, J.~Xie, and X.~Li, ``Visual haze removal by a unified generative
  adversarial network,'' \emph{IEEE Transactions on Circuits and Systems for
  Video Technology}, vol.~29, no.~11, pp. 3211--3221, 2018.

\bibitem{zhao2020gradient}
H.~Zhao, D.~Wu, H.~Su, S.~Zheng, and J.~Chen, ``Gradient-based conditional
  generative adversarial network for non-uniform blind deblurring via
  denseresnet,'' \emph{Journal of Visual Communication and Image
  Representation}, p. 102921, 2020.

\bibitem{dr-gan}
K.~Liao, C.~Lin, Y.~Zhao, and M.~Gabbouj, ``Dr-gan: Automatic radial distortion
  rectification using conditional gan in real-time,'' \emph{IEEE Transactions
  on Circuits and Systems for Video Technology}, 2019.

\bibitem{li2020improved}
C.~Li, L.~Kong, and Z.~Zhou, ``Improved-storygan for sequential images
  visualization,'' \emph{Journal of Visual Communication and Image
  Representation}, vol.~73, p. 102956, 2020.

\bibitem{fu2020conditional}
B.~Fu, F.~Li, Y.~Niu, H.~Wu, Y.~Li, and G.~Shi, ``Conditional generative
  adversarial network for eeg-based emotion fine-grained estimation and
  visualization,'' \emph{Journal of Visual Communication and Image
  Representation}, p. 102982, 2020.

\bibitem{zhang2020semi}
L.~Zhang, L.~Chen, W.~Ou, and C.~Zhou, ``Semi-supervised cross-modal
  representation learning with gan-based asymmetric transfer network,''
  \emph{Journal of Visual Communication and Image Representation}, vol.~73, p.
  102899, 2020.

\bibitem{grvkgan}
S.~Wen, W.~Liu, Y.~Yang, T.~Huang, and Z.~Zeng, ``Generating realistic videos
  from keyframes with concatenated gans,'' \emph{IEEE Transactions on Circuits
  and Systems for Video Technology}, vol.~29, no.~8, pp. 2337--2348, 2018.

\bibitem{cui2020ap}
R.~Cui, G.~Hua, and J.~Wu, ``Ap-gan: Predicting skeletal activity to improve
  early activity recognition,'' \emph{Journal of Visual Communication and Image
  Representation}, vol.~73, p. 102923, 2020.

\bibitem{chuk}
X.~Wang and X.~Tang, ``Face photo-sketch synthesis and recognition,''
  \emph{IEEE Transactions on Pattern Analysis \& Machine Intelligence}, no.~11,
  pp. 1955--1967, 2008.

\bibitem{facades}
R.~Tyle{\v c}ek and R.~{\v S}{\' a}ra, ``Spatial pattern templates for
  recognition of objects with regular structure,'' in \emph{German Conference
  on Pattern Recognition}, 2013.

\bibitem{rgb_nir}
M.~Brown and S.~S\"usstrunk, ``Multispectral {SIFT} for scene category
  recognition,'' in \emph{Computer Vision and Pattern Recognition (CVPR11)},
  Colorado Springs, June 2011, pp. 177--184.

\bibitem{dualGAN}
Z.~Yi, H.~Zhang, P.~Tan, and M.~Gong, ``Dualgan: Unsupervised dual learning for
  image-to-image translation,'' in \emph{IEEE International Conference on
  Computer Vision}, 2017, pp. 2868--2876.

\bibitem{CyclicGAN}
J.-Y. Zhu, T.~Park, P.~Isola, and A.~A. Efros, ``Unpaired image-to-image
  translation using cycle-consistent adversarial networks,'' in \emph{IEEE
  International Conference on Computer Vision}, 2017, pp. 2242--2251.

\bibitem{FCNSS}
J.~Long, E.~Shelhamer, and T.~Darrell, ``Fully convolutional networks for
  semantic segmentation,'' in \emph{Proceedings of the IEEE conference on
  computer vision and pattern recognition}, 2015, pp. 3431--3440.

\bibitem{AC}
G.~Larsson, M.~Maire, and G.~Shakhnarovich, ``Learning representations for
  automatic colorization,'' in \emph{European Conference on Computer
  Vision}.\hskip 1em plus 0.5em minus 0.4em\relax Springer, 2016, pp. 577--593.

\bibitem{fccn}
L.~Zhang, L.~Lin, X.~Wu, S.~Ding, and L.~Zhang, ``End-to-end photo-sketch
  generation via fully convolutional representation learning,'' in \emph{ACM
  International Conference on Multimedia Retrieval}, 2015, pp. 627--634.

\bibitem{ISTCNN}
L.~A. Gatys, A.~S. Ecker, and M.~Bethge, ``Image style transfer using
  convolutional neural networks,'' in \emph{Proceedings of the IEEE Conference
  on Computer Vision and Pattern Recognition}, 2016, pp. 2414--2423.

\bibitem{feng2018dual}
Z.~Feng, X.~Wang, C.~Ke, A.~Zeng, D.~Tao, and M.~Song, ``Dual swap
  disentangling,'' in \emph{NeurIPS}, 2018.

\bibitem{gan}
I.~Goodfellow, J.~Pouget-Abadie, M.~Mirza, B.~Xu, D.~Warde-Farley, S.~Ozair,
  A.~Courville, and Y.~Bengio, ``Generative adversarial nets,'' in
  \emph{Advances in neural information processing systems}, 2014, pp.
  2672--2680.

\bibitem{PRSISRGAN}
C.~Ledig, L.~Theis, F.~Husz{\'a}r, J.~Caballero, A.~Cunningham, A.~Acosta,
  A.~P. Aitken, A.~Tejani, J.~Totz, Z.~Wang \emph{et~al.}, ``Photo-realistic
  single image super-resolution using a generative adversarial network.'' in
  \emph{CVPR}, vol.~2, no.~3, 2017, p.~4.

\bibitem{FAGSPS}
H.~Kazemi, M.~Iranmanesh, A.~Dabouei, S.~Soleymani, and N.~M. Nasrabadi,
  ``Facial attributes guided deep sketch-to-photo synthesis,'' in
  \emph{Computer Vision Workshops (WACVW), 2018 IEEE Winter Applications
  of}.\hskip 1em plus 0.5em minus 0.4em\relax IEEE, 2018, pp. 1--8.

\bibitem{bridge-gan}
M.~Yuan and Y.~Peng, ``Bridge-gan: Interpretable representation learning for
  text-to-image synthesis,'' \emph{IEEE Transactions on Circuits and Systems
  for Video Technology}, 2019.

\bibitem{ddr3dcgan}
Y.~Hu, M.~Lu, C.~Xie, and X.~Lu, ``Driver drowsiness recognition via 3d
  conditional gan and two-level attention bi-lstm,'' \emph{IEEE Transactions on
  Circuits and Systems for Video Technology}, 2019.

\bibitem{RS-DAN}
C.~Wang, M.~Niepert, and H.~Li, ``Recsys-dan: Discriminative adversarial
  networks for cross-domain recommender systems,'' \emph{IEEE transactions on
  neural networks and learning systems}, 2019.

\bibitem{conditionalGAN}
M.~Mirza and S.~Osindero, ``Conditional generative adversarial nets,''
  \emph{arXiv preprint arXiv:1411.1784}, 2014.

\bibitem{cGAN}
P.~Isola, J.-Y. Zhu, T.~Zhou, and A.~A. Efros, ``Image-to-image translation
  with conditional adversarial networks,'' in \emph{IEEE Conference on Computer
  Vision and Pattern Recognition}, 2017, pp. 5967--5976.

\bibitem{PAN}
C.~Wang, C.~Xu, C.~Wang, and D.~Tao, ``Perceptual adversarial networks for
  image-to-image transformation,'' \emph{IEEE Transactions on Image
  Processing}, vol.~27, no.~8, pp. 4066--4079, 2018.

\bibitem{ps2-man}
L.~Wang, V.~Sindagi, and V.~Patel, ``High-quality facial photo-sketch synthesis
  using multi-adversarial networks,'' in \emph{IEEE International Conference on
  Automatic Face \& Gesture Recognition}, 2018, pp. 83--90.

\bibitem{csgan}
K.~B. Kancharagunta and S.~R. Dubey, ``Csgan: Cyclic-synthesized generative
  adversarial networks for image-to-image transformation,'' \emph{arXiv
  preprint arXiv:1901.03554}, 2019.

\bibitem{LSGAN}
X.~Mao, Q.~Li, H.~Xie, R.~Y. Lau, Z.~Wang, and S.~P. Smolley, ``Least squares
  generative adversarial networks,'' in \emph{IEEE International Conference on
  Computer Vision}, 2017, pp. 2813--2821.

\bibitem{adam}
D.~P. Kingma and J.~Ba, ``Adam: A method for stochastic optimization,''
  \emph{International Conference on Learning Representations}, 2014.

\bibitem{dcgan}
A.~Radford, L.~Metz, and S.~Chintala, ``Unsupervised representation learning
  with deep convolutional generative adversarial networks,'' \emph{arXiv
  preprint arXiv:1511.06434}, 2015.

\bibitem{SSIM}
Z.~Wang, A.~C. Bovik, H.~R. Sheikh, and E.~P. Simoncelli, ``Image quality
  assessment: from error visibility to structural similarity,'' \emph{IEEE
  transactions on image processing}, vol.~13, no.~4, pp. 600--612, 2004.

\bibitem{LPIPS}
R.~Zhang, A.~A. Efros, E.~Shechtman, and O.~Wang, ``The unreasonable
  effectiveness of deep features as a perceptual metric,'' in \emph{IEEE
  International Conference on Computer Vision}, 2018.

\end{thebibliography}

\end{document}